\documentclass{article}
\usepackage{spconf}
\usepackage{graphicx}
\usepackage[cmex10]{amsmath}
\usepackage{amsfonts}
\usepackage{xcolor}
\usepackage{amsfonts}
\usepackage{amssymb}
\usepackage{amsthm}
\usepackage{booktabs}
\usepackage{algorithm}
\usepackage{algorithmic}
\usepackage[misc]{ifsym}
\usepackage{times}

\usepackage{soul}
\usepackage{dblfloatfix}    
\usepackage{url}
\usepackage{xcolor}
\usepackage{float}
\usepackage{cite}
\usepackage[pagebackref=false,breaklinks=true,colorlinks,bookmarks=false]{hyperref}
\usepackage[caption=true,font=normalsize,labelfont=sf,textfont=sf]{subfig}

\DeclareMathOperator*{\argmin}{arg\,min}
\hypersetup{pdfinfo={Title={Bringing Computer Vision closer to Cognitive Psychology: Teaching CNNs to mimic Human Visual Cognitive Process \& regularise Texture-Shape Bias},
    Author={Satyam Mohla, Anshul Nasery, Biplab Banerjee},
    Subject={Computer Vision, Image Processing, Explainable, Interpretable, AI, CNN, Experimental, bias},
    Keywords={feature, image, texture, network, shape, attention, edges, given, stream, cognitive, reconstruction, experiments, robustness, latent, greyscale, classify, treisman, interpretable}}}
\begin{document}
\title{Teaching CNNs to mimic Human Visual Cognitive \\ Process \& regularise Texture-Shape Bias}
%

\name{Satyam~Mohla$^{\dagger}$\href{https://orcid.org/0000-0002-5400-1127}{\protect \includegraphics[scale=0.12]{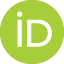}}, Anshul~Nasery, Biplab~Banerjee \href{https://orcid.org/0000-0001-8371-8138}{\protect \includegraphics[scale=0.12]{images/orcid.png}\vspace{-10pt}}
\thanks{\hspace{-10pt}$^{\dagger}$Corresponding Author. Currently at DTSU Honda Innovation Lab Tokyo.}
\address{Indian Institute of Technology, Bombay, India\\
\protect\includegraphics[scale=0.15]{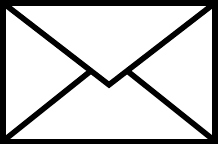}\href{mailto:satyammohla@iitb.ac.in}{\hspace{3pt}satyammohla@iitb.ac.in}}}
\vspace{-10pt}
%
    
\maketitle

\begin{abstract}{
Recent experiments in computer vision demonstrate texture bias as the primary reason for supreme results in models employing Convolutional Neural Networks (CNNs), conflicting with early works claiming that these networks identify objects using shape. It is believed that the cost function forces the CNN to take a greedy approach and develop a proclivity for local information like texture to increase accuracy, thus failing to explore any global statistics. We propose CognitiveCNN, a new intuitive architecture, inspired from feature integration theory in psychology to utilise human-interpretable feature like shape, texture, edges etc. to reconstruct, and classify the image. We define novel metrics to quantify the "relevance" of "abstract information" present in these modalities using attention maps. We further introduce a regularisation method which ensures that each modality like shape, texture etc. gets proportionate influence in a given task, as it does for reconstruction; and perform experiments to show the resulting boost in accuracy and robustness, besides imparting explainability to these CNNs for achieving superior performance in object recognition.
}
\end{abstract}
\vspace{-5pt}
\section{Introduction}
\label{sec:intro}
\vspace{-5pt}
{CNNs, considered as the computational model for primate visual system\cite{cadieu2014deep,kubilius2016deep}, have been shown to exhibit representation hierarchy in terms of feature selectivities of edges, shapes and objects in early, mid and deep level units. The fact that complex objects and shapes appear after edges in intermediate layer activation visualisations of CNNs seem to support empirically a theoretical understanding of interpretable selectivities\cite{kriegeskorte2015deep, gucclu2015deep}, also in agreement with the shape bias observed in experiments with children\cite{ritter2017cognitive}.

However, recent works demonstrate texture bias as the reason for the superior performance of CNNs\cite{imagenettexture}. Similar conclusions were drawn in\cite{brendel2019approximating}, where texturised images of dogs were correctly classified, even when global statistics were highly distorted. It seems that CNNs, in order to maximise accuracy, greedily learned to use texture to solve the problem and thus failed to learn the global features relevant for the task. \cite{imagenettexture} attempts to reduce this bias by training an ImageNet pretrained CNN with a stylised texture image dataset. The method although novel, is ad-hoc and does not address the underlying problem of greedy learning in CNNs. Moreover, such techniques are difficult to apply in tasks where any non-domain training data can lead to possible distortion and thus to loss in accuracy \& robustness, or where the image data is inherently of low quality. 

In this paper,we propose methods to utilise self-attention for quantifying the texture-shape bias in CNNs in an intuitive \& interpretable manner. Furthermore, we introduce new metrics to regulate bias \& demonstrate resulting gains in performance \& robustness in object recognition.}
  \vspace{-5pt}
\section{Feature Integration Theory (FIT):}
  \vspace{-5pt}
\begin{figure}[t]%
  \centering
  \centerline{\includegraphics[width=8cm]{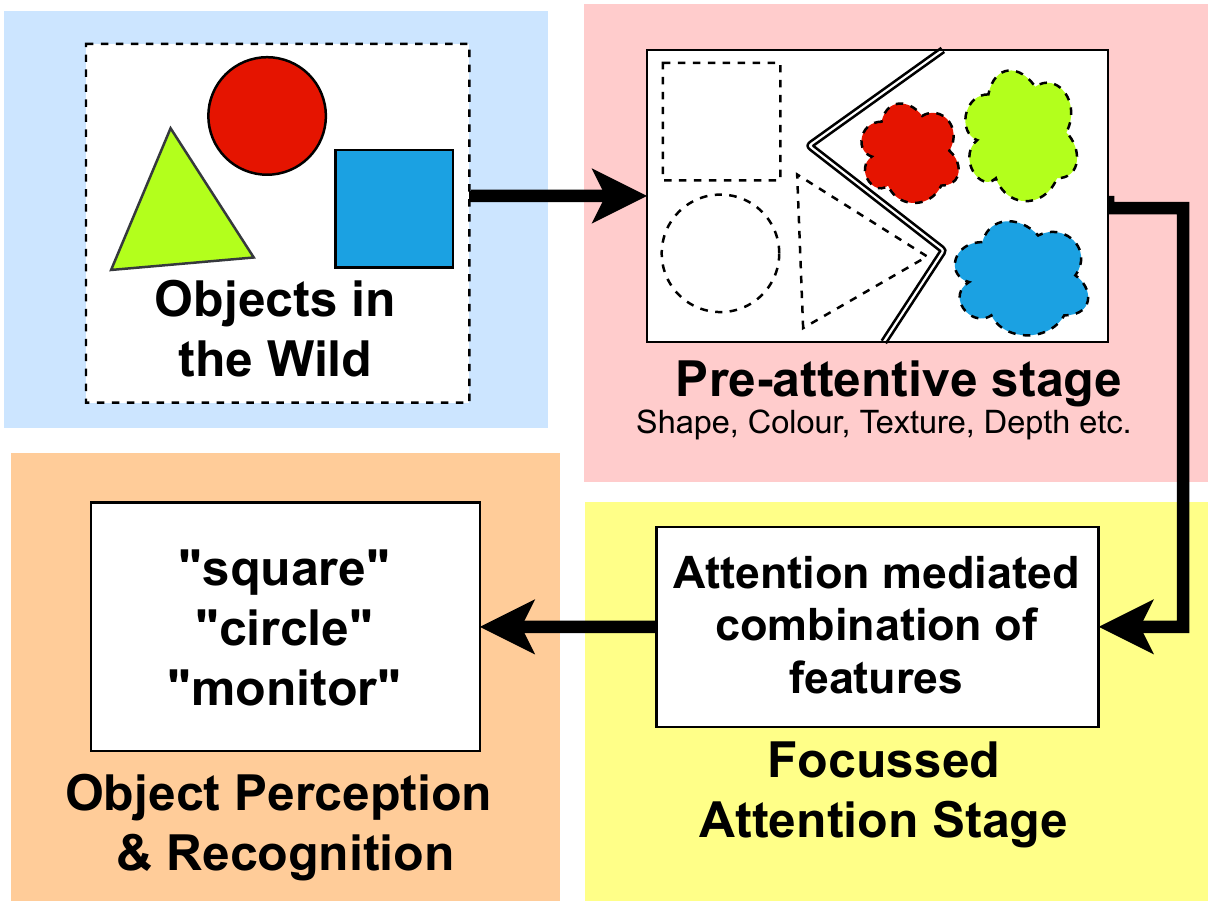}}
  \caption{Stages of FIT based object recognition.}
  \label{treisman}
\end{figure}

In cognitive psychology, FIT refers to an attention model which suggests that when perceiving objects, we first synthesise and separate features in an automatic \& parallel way, directing attention serially to each item in turn afterwards\cite{Treisman1980-TREAFI}, as shown in Figure \ref{treisman}. This has been supported by many experiments\cite{treisman1982illusory, friedman1995parietal,robertson1997oops}.The features isolated in pre-attentive stage include shape, colour, size, curvature, lines etc. \cite{treisman1986features}.

\begin{figure*}[t] 
  \centering
  \centerline{\includegraphics[width=17cm]{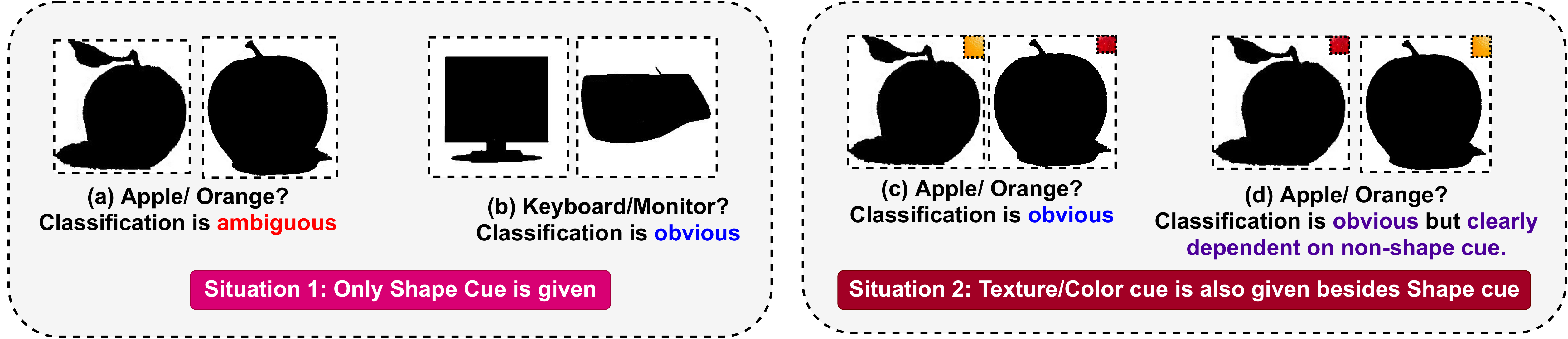}}
  \vspace{-5pt}
  \caption{Gedankenexperiment: Importance of Shape Cue vs Shape + Texture cue for varying classes}\medskip
  \label{pstatement}
  \vspace{-5pt}
\end{figure*}
\begin{figure*}[!b] 
  \centering
  \centerline{\includegraphics[width=19cm]{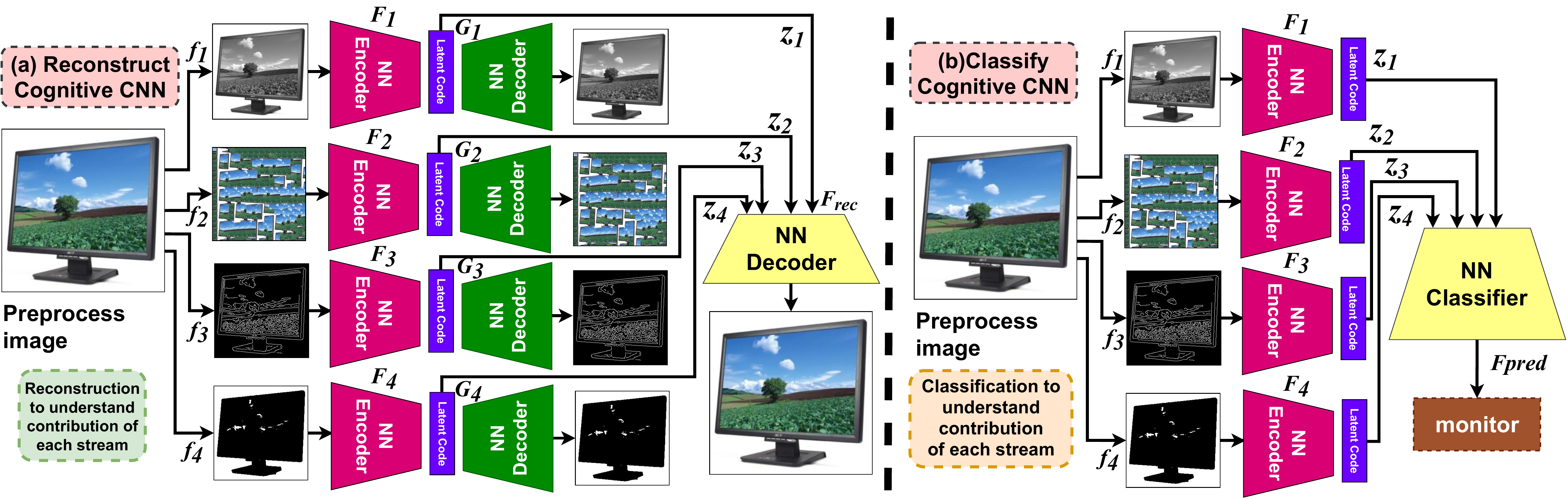}}
  \vspace{-0.3 cm}
  \caption{Experimental Setup: FIT model adapted to CNN for quantification and regularisation}\medskip
  \label{architecture}  
  \label{ret}
  \vspace{-0.5cm}
\end{figure*}
FIT provides a novel inspiration to combat our problem: we provide different feature selectivities as input to the network, emulating the pre-attentive stage. Our model can thus train with the knowledge of various features like texture, shape and edges and explore more avenues to maximise performance.
\vspace{-10pt}

\subsection{Gedankenexperiment}
Let us begin with a thought experiment. Consider humans tasked with classifying a pair of silhouette of objects, for example, orange/apple \& keyboard/monitor as in Fig \ref{pstatement}. In situation 1, where only shape cue is given, the classes are somewhat ambiguous in \ref{pstatement}(a), but obvious in \ref{pstatement}(b). In situation 2 however, where additional information of texture/colour is given, a human can easily classify orange/apple.

Note that in case of classifying orange/apple silhouette, no information was imparted by the silhouette (shape) to human. It was the texture/colour cue that assisted us make the decision. To confirm, consider \ref{pstatement}(d), which has opposite texture cue as to \ref{pstatement}(c). Does the result of your classification reverses when the texture cue reverses too?

Essentially the CNN, just like humans, should learn to give different relevance to different modalities of information depending on the task. Feature Integration Theory (FIT) has been experimentally observed in various studies \& thus provides the inspiration for our investigation.

\vspace{-10pt}
\section{Experimental Setup}
\vspace{-5pt}
\subsection{Preprocessing}
Let the original dataset be $ \{ x,y \}^{m}_{i=1}$ where $x$ is an image and $y$ is the associated label. Further, let $f_1,f_2,...,f_n$ be a set of \textit{feature transform based classical image processing algorithm} which can be applied in preprocessing stage to each $x$ to extract a new modality. In our case, $f_1$, $f_2$, $f_3$, and $f_4$ are instantiated to extract shape, texture, greyscale image and edges respectively, \& each resulting $f_i(x)$ is a new modality. Additional modalities are generated similarly as in \cite{imagenettexture}:

\textbf{Greyscale:} \hspace{0.2cm} Images are processed in Matlab and converted to greyscale using \texttt{skimage.color.rgb2gray}

\textbf{Silhouette:} \hspace{0.2cm} These are generated by thresholding colour images on white background. As such the outermost contour is interpreted as the "perceived" shape, which is how human perception would also interpret the object when looked.

\textbf{Edges} \hspace{0.2cm} The edge representation for the image is generated in Matlab using \texttt{edge(I, `Canny')}

\begin{figure}[!b] 
  \centering
  \centerline{\includegraphics[width=8cm]{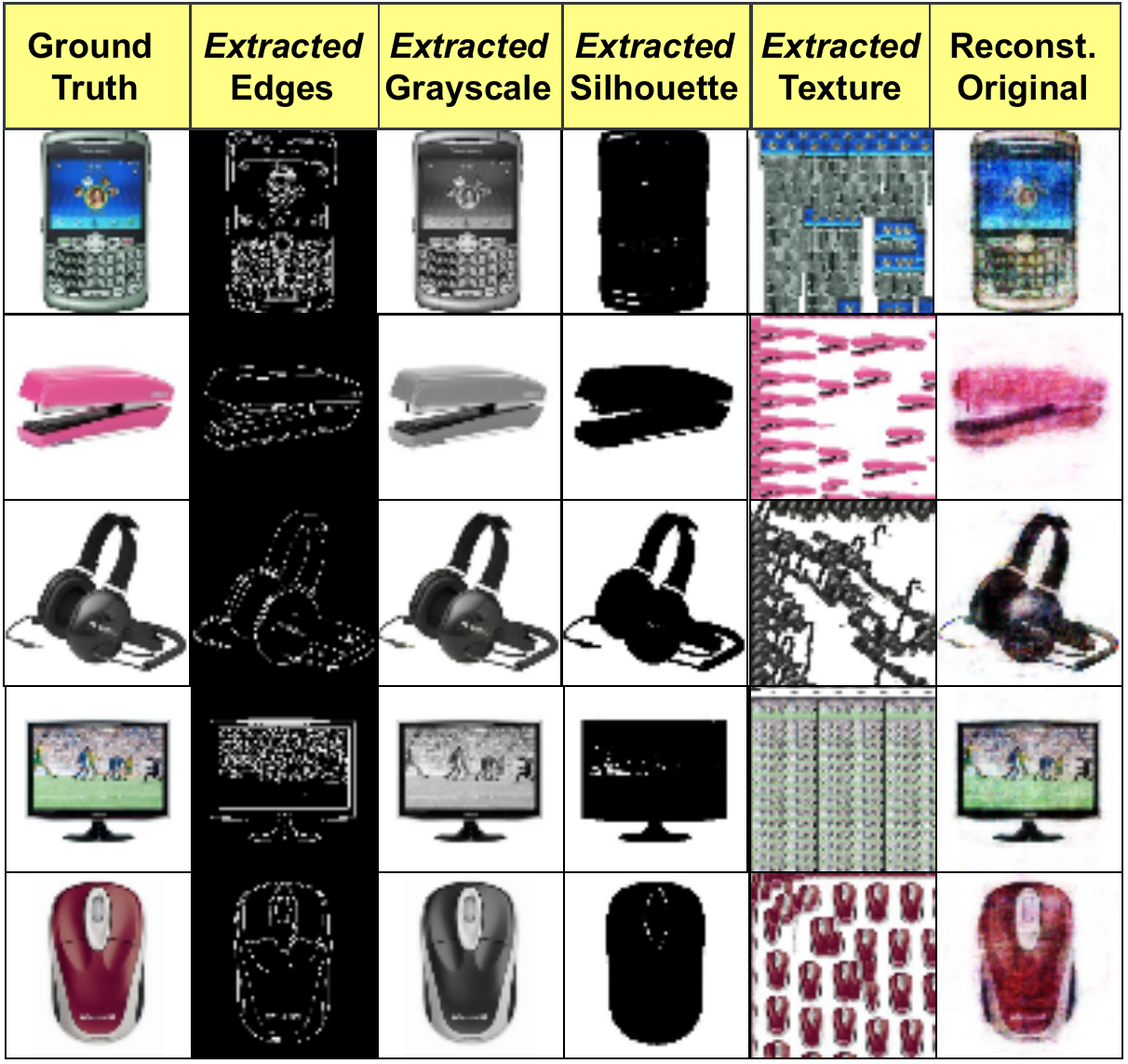}}
    \caption{Some examples of reconstructions produced by our Reconstruct Cognitive Network}
\medskip
  \label{reconstruction}
\end{figure}

\textbf{Texture} \hspace{0.2cm} We utilise the interpretation of \cite{imagenettexture} to define texture as repetition: Many repeated `things' become `stuff'\cite{gatys2017texture,imagenettexture}. 
We utilise \cite{efros2001image}, to generate texture classically (to ensure deterministic \& reproducible outputs) as shown in Fig\ref{reconstruction}.

\vspace{-5pt}

\vspace{-5pt}
\subsection{Architecture}
\vspace{-5pt}
Next, we describe our model's architecture and training method. We represent our model by the tuple $(F_i(x)^{i=4}_{i=1},$ $F_{rec}, F_{pred})$ where each $F_i(x)^{i=4}_{i=1}$ acts as a modality encoder for corresponding $f_i(x)$, converting it to a latent vector $z_i$. $F_{rec}$ represents the network tasked with reconstructing the original image, and $F_{pred}$ represents the network for predicting the labels for a given set of input modality tuple. $G_i$ are decoders tasked to assist encoders $F_i$ to learn the latent representation of modality in autoencoder setting, as in Fig \ref{architecture}.

For the modality encoder $(F_i(\theta_i))$, we used a SegNet  backbone\cite{segnet}, and for generating attention maps (described in Section \ref{section:attention}), we used a 2-layer Conv Network (with 128 filters, kernel size of 2) taking the entire concatenated latent vector(outputs of block 5 of SegNet) as input \& generating attended vectors as output. Lastly, to predict the label, we feed these attended vectors through a BatchNorm \& Dense layer with softmax activation.

\vspace{-5pt}
\subsection{Training}
In the first stage of training, we train our modality encoders $F_i(\theta_i)$ with learnable parameters $\theta_{i}$ and reconstruction network $F_{rec}$ to reconstruct the original image from the given modality encoders. The input to $F_{rec}$ are the concatenated latent vectors $z_1,z_2,...,z_n$ (Figure \ref{architecture} (a)). The purpose of this part of training is to tune the modality encoders, and to gauge the "relevance" of each modality in the reconstruction for the image. The reconstruction also confirms the assumption that all the information of the image is captured in these four modalities. Formally, this stage of training can be summarized as in equation (\ref{eqn1}):
\begin{equation}
\begin{split}
    \argmin_{\theta_{1}, \theta_{2},...,\theta_{n}, \theta_{rec}} \mathcal{D}\biggl(F_{rec}\Bigl(F_1\bigl(f_1(x)\bigr), F_2\bigl(f_2(x)\bigr),...
    \\
...,F_n\bigl(f_n(x)\bigr)\Bigr),x\biggr) + \lambda\sum_{i=1}^n\mathcal{D}\biggl(G_i\bigl(F_i\bigl(f_i(x)\bigr)\bigr), f_i(x)\biggr)
\end{split}
\label{eqn1}
\end{equation}
where $\mathcal{D}$ represents the pixel-wise Euclidean distance between original and reconstructed images. \\
Once the networks have converged, we train the prediction network $F_{pred}$ to predict the label of each input given its latent vectors $z_1,z_2,...,z_n$. (Figure \ref{architecture} (b)). Formally, this stage aims to find $F_{pred}(\theta_{pred})$ as in equation (\ref{eqn2}):
\begin{equation}
\begin{split}
    F_{pred}(\theta_{pred}) \leftarrow  \argmin_{(\theta_{pred})} \mathcal{L}_{ce}\biggl(F_{pred}\Bigl(F_1\bigl(f_1(x)\bigr),
    \\
    F_2\bigl(f_2(x)\bigr), F_3\bigl(f_3(x)\bigr),...,F_n\bigl(f_n(x)\bigr)\Bigr),y\biggr)
\end{split}
\label{eqn2}
\end{equation}
where $\mathcal{L}_{ce}$ is the cross-entropy loss between the predicted labels and true labels. Our setup is summarized in Fig.\ref{architecture}.\\
Now that we have classified or generated the image from the human interpretable features, we want to quantify the "relative relevance"  among them for the different tasks, for which we use attention maps.

\vspace{-10pt}
\section{Quantifying Bias using Attention}
\label{section:attention}
In this section, we introduce a self-attention based mechanism to quantify the bias in the dataset as well as our prediction network. There have been previous attempts to use attention as a tool for neural feature selection (\cite{wang2014attentional}, \cite{gui2019afs}). We extend this technique to utilize attention as a means to weigh the relative relevance of each modality in prediction and reconstruction, and to further regulate information flow to prediction network in an unbiased manner to make it more robust.

Let $A^{j=4}_{j=1}$ be the attention layers corresponding to the modalities. We add self-attention layers $A_{pred}$ and $A_{rec}$ to the network which act on the concatenated latent vectors $z_1,...,z_n$ to give weighted vectors $\hat{z}_1,..., \hat{z}_n$. These are then passed to $F_{pred}$ and $F_{rec}$ to classify and reconstruct respectively. Formally, 
\begin{equation*}
    \hat{z} = \sigma(A_j(z)) \odot z,  \qquad z = (z_1 \mathbin\Vert z_2\mathbin\Vert,...,\mathbin\Vert z_n)
\end{equation*}
where $\odot$ represents element wise product, $z$ is the concatenated latent vector.

\vspace{-10pt}
\subsection{Measuring Shape \& Texture Cue}
\vspace{-5pt}
It is these attention maps we use to quantify the biases. We define the measures \textit{Reconstruction Relevant Modality Coefficient (RRMC)} and \textit{Task Relevant Modality Coefficient (TRMC)} for each modality $i$ for a particular example as 
\begin{align*}
    \textit{RRMC}_i(z)&= \frac{\mathbb{E}(\sigma(A_{rec}(z))_i)}{\sum_{j=1}^n \mathbb{E}( \sigma(A_{rec}(z))_j)} \
    \end{align*}
    \vspace{-10pt}
    \begin{align*}
    \textit{TRMC}_i(z)  = \frac{\mathbb{E}(\sigma(A_{pred}(z))_i)}{\sum_{j=1}^n \mathbb{E}(\sigma(A_{pred}(z))_j)}    
\end{align*}
where $\mathbb{E}$ represents the mean of a  vector over its dimensions. 
\textit{RRMC$_i$} and \textit{TRMC$_i$} represent the abstract measure of amount of "relative relevance" of a given modality $i$ for reconstruction and prediction networks respectively. This "relative relevance" is reflected in the attention maps that the network generates, assigning maximal importance to that modality that assists it the most; to reconstruct the image in $\textit{RRMC}_{i}$, \& towards classifying the given image in $\textit{TRMC}_{i}$. Finally, prediction network is defined as biased if there is a mismatch in this "relative relevance" of the modality for the two tasks, namely reconstruction and prediction. i.e. when $\textit{TRMC}_{i}$ is not equal to ${RRMC}_{i}$
\vspace{-10pt}

%

\subsection{Regularising Shape-Texture Bias}
\label{section:regulariser}
To control the shape-texture bias, we add a regularizer $\sum_{i=1}^{n=4} ||\textit{TRMC}_i - \textit{RRMC}_i||^2$ to the loss function, which forces the prediction network to give as much importance ($\textit{TRMC}_{i}$) to a modality for a given task (herein the task is prediction), as much as it was important ($\textit{RRMC}_{i}$) for reconstruction.

\begin{table}[b]
\vspace{15pt}\caption{Comparing relative relevance of different modalities}
\vspace{-15pt}
\label{recdiff}
\begin{center}
\begin{tabular}{llll}
\hline
\textbf{Stream} & \textbf{$\textit{RRMC}_i$} & {\textbf{$\textit{TRMC}_i$}}& {\textbf{$\textit{TRMC}_i$}}
\\
{} & {} & \textbf{4UC-CogCNN} & \textbf{4RC-CogCNN}\\ \hline 
Shape         & $23.7\%$ & $21.0\%$ & $24.0\%$ \\
Texture       & $22.3\%$ & $22.8\%$ &  $22.2\%$ \\
Greyscale     & $30.4\%$ & $31.4\%$ & $30.7\%$ \\
Edges         & $23.4\%$ & $24.6\%$ & $23.0\%$ \\
\hline
{\textbf{Accuracy}}    & & $58.7\%$ & $61.8\%$ \\
\hline
\end{tabular}
\end{center}
\end{table}
\section{Results}

Since our tasks involve  pre-processing using classical techniques, we utilised a dataset with a white background, like Amazon Office-31 dataset\cite{saenko2010adapting}. We perform experiments \& show the efficacy of our measures \& regularizer, and demonstrate gains in performance and robustness due to our method.
\vspace{-18pt}

\subsection{Reconstructions}
The first stage of training involves reconstructing the image from the four modalities. The reconstructions (Figure \ref{reconstruction}) look very close to the original image,
which demonstrates that (i) the chosen modalities contained all information of the original image and (ii) the current setup was able to extract all the information from these modalities into the latent code, essentially \emph{demonstrating an empirical scheme to extract numerical representation of abstract modalities like shape, texture, edge cues etc.}, basically what we set out to achieve.

\vspace{-10pt}
\subsection{Accuracy}
We trained a network using our method (Figure \ref{architecture}) to classify the Amazon Office-31 dataset and recorded the value of $\textit{RRMC}_i$ and $\textit{TRMC}_i$ for each feature, besides the accuracy. Unregularised network (4UC-CogCNN) reported an accuracy of $58.7\%$, while the accuracy in network (4RC-CogCNN) resulting due to regularisation (Section \ref{section:regulariser}) increased to $61.8\%$, shown in Table \ref{recdiff}.

We incorporate these ideas into our baseline CNN, and propose Cue Augmented CNN (CueAugCNN) which takes all 4 features as additional input channels alongside the image itself. We compared all the methods with a baseline CNN having the same architecture. All our 4 stream networks (4UC, 4RC, CueAugCNN) perform superior to the baseline network, with the highest accuracy being achieved by CueAugCNN at $62.5\%$. The results are shown in Table \ref{accrob}.

\begin{figure}[t!] 
  \centering
  \centerline{\includegraphics[width=8.7cm]{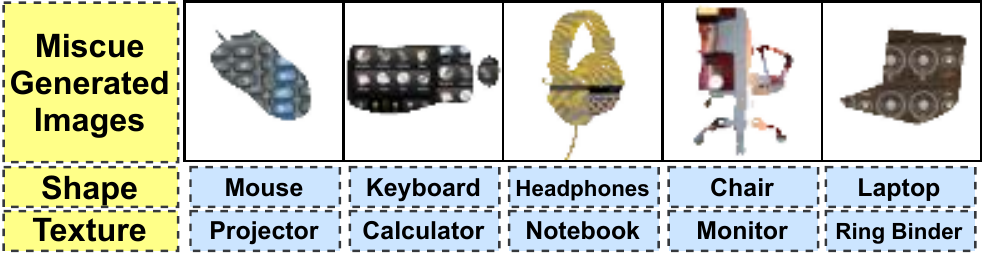}}
\caption{Examples of Miscue Conflict}
\label{miscueimg}
\end{figure}

\subsection{Robustness}
To test for robustness, we performed a texture-shape miscue experiment as done in \cite{imagenettexture}. In order to demonstrate miscue conflict, we classically generated images by overlaying the texture of one object on the shape of an object from another class, as shown in Fig \ref{miscueimg}.


All CogCNN approaches performed consistently better in robustness than the baseline by a large margin. CueAugCNN however, based on conventional CNN architecture performed poorly, at the cost of increase in accuracy. In an ablation of CogCNN, we considered only 2 streams (shape-texture). The network still performed comparable to baseline (only 0.7\% decrease in accuracy) for a huge gain in \mbox{robustness. Our results are tabulated in Table \ref{accrob}.}
\vspace{12pt}
\begin{table}[htb]
\caption{Accuracy \& robustness for different models}
\vspace{-12pt}
\label{accrob}
\begin{center}
\begin{tabular}{cll}
\hline
\textbf{Method} &
\multicolumn{2}{c}{\textbf{Accuracy}} \\
{} & \textbf{Original} & \textbf{Miscue} \\
\hline 
Conventional CNN (Baseline)        & $58.3\%$ & $14.5\%$  \\
2 Stream Reg (2RC-CogCNN) & $57.6\%$ & ${49.3}\% $  \\
4 Stream Unreg (4UC-CogCNN) & ${58.7\%}$ & ${52.0\%}$  \\
4 Stream Reg (4RC-CogCNN) & $\textbf{61.8\%}$ & $\textbf{56.9}$\%  \\
CueAugmented (CueAugCNN)         & $\textbf{62.5\%}$ & $11.1\%$ \\
\hline
\end{tabular}
\end{center}
\end{table}

\vspace{-30pt}
\section{Conclusion}
\vspace{-10pt}
We demonstrated an empirical scheme to extract numerical representation of abstract modalities like shape, texture, edge cues etc., imparting explainability to the model in tasks like reconstruction \& classification. We developed novel metrics \& regulariser to control bias between different modalities, like texture-shape bias in the network. We showed that training a CNN with human-interpretable modalities like shape/texture/edge cues, as inspired from FIT, lead to increase in accuracy \& robustness against cue conflicts. Lastly, we adapted the ideas to conventional CNNs, and achieved highest accuracy. Our future work includes preprocessing the input image in-situ to present an end-to-end network so that it can be readily used on any dataset.
\bibliographystyle{IEEEbib}
{\bibliography{root}}
\end{document}